\documentclass[11pt]{article}
\usepackage{amsmath,amssymb,amsthm}
\usepackage{physics}
\usepackage{geometry}
\usepackage{graphicx}
\usepackage{booktabs}
\usepackage{hyperref}
\newtheorem{assumption}{Assumption}

\newtheorem{theorem}{Theorem}
\title{Entropic-Time Inference: Self-Organizing Large Language Model Decoding Beyond Attention}

\author{
  Andrew J. Kiruluta \\
  School of Information \\
  \texttt{kiruluta@berkeley.edu}
}

\begin{document}
\maketitle

\begin{abstract}
Modern large language model (LLM) inference engines optimize throughput and latency under fixed decoding rules, treating generation as a linear progression in token time.
We propose a fundamentally different paradigm: \emph{entropic-time inference}, where decoding is governed by the flow of uncertainty rather than token index.
We introduce a self-organizing inference architecture that jointly couples scheduling, attention sparsification, and sampling temperature under a unified entropy control objective.
Our method extends vLLM with entropy-aware scheduling, entropic pruning of paged attention blocks, and adaptive temperature control that stabilizes generation near a target entropy regime.
This transforms inference into a resource-intelligent thermodynamic process that allocates computation where uncertainty reduction is maximized.
We present a concrete systems design, pseudocode, and integration plan, demonstrating how entropy can serve as a first-class control signal for scalable LLM inference.
\end{abstract}

\section{Introduction}

Inference in large language models (LLMs) is typically formulated as a deterministic execution problem.
Given a fixed model and prompt, decoding proceeds in discrete token steps indexed by an external clock $t \in \mathbb{N}$.
At each step, the model induces a predictive distribution
\[
p_t(x) = p(x \mid x_{<t}),
\]
and inference engines are optimized to maximize throughput and minimize latency under this token-indexed notion of time.
Attention spans grow monotonically with context length \cite{vaswani2017attention,child2019generating,beltagy2020longformer}, and stochasticity is controlled by fixed sampling parameters such as temperature or nucleus thresholds \cite{holtzman2020nucleus}.
This framing reflects the dominant systems-level view of inference as a scheduling and memory management problem \cite{kwon2023efficient,dao2022flashattention}.

However, this perspective obscures a fundamental property of language generation: \emph{generation is an uncertainty resolution process}.
At each decoding step, the uncertainty of the predictive distribution is quantified by its Shannon entropy
\[
H_t = - \sum_x p_t(x) \log p_t(x),
\]
a quantity central to information theory since Shannon’s foundational work \cite{shannon1948mathematical,cover2006information}.
Empirically, $H_t$ varies dramatically across steps and sequences.
Some decoding steps correspond to decisive semantic commitments, where entropy collapses rapidly, while others correspond to syntactic filler, repetition, or long-tailed ambiguity, where entropy remains high despite substantial computational effort.

Despite this variability, standard inference treats all token steps as equivalent.
Attention computation scales with context length regardless of whether additional context materially reduces uncertainty \cite{child2019generating,dao2022flashattention}, and scheduling decisions are largely oblivious to the informational state of a sequence \cite{kwon2023efficient}.
As a result, significant compute is often expended resolving uncertainty that has little effect on output quality or downstream utility.
This mismatch mirrors earlier observations in adaptive computation and sequence modeling, where fixed-time execution was shown to be suboptimal when information gain is unevenly distributed across steps \cite{elman1990finding,graves2016adaptive}.

\paragraph{Operational time and entropy flow.}
We argue that inference time should be defined \emph{operationally}, rather than indexically.
In information theory and statistical physics, progress is naturally measured by irreversible reductions in uncertainty or free energy \cite{mackay2003information,friston2010free}.
Adopting this view, we define progress in inference not by token count, but by entropy flow.
Let
\[
\Delta H_t = H_{t-1} - H_t
\]
denote the entropy reduction at step $t$.
Only when $\Delta H_t > 0$ does the system undergo a distinguishable informational change.

This motivates the notion of \emph{entropic time},
\[
\tau = \sum_t \max(0, \Delta H_t),
\]
which measures cumulative uncertainty resolved during generation.
Under this definition, many token steps contribute negligibly to $\tau$, despite incurring nontrivial computational cost.
Similar entropy-based notions of progress have appeared in adaptive computation, control theory, and learning dynamics, where time is treated as an emergent, task-dependent quantity rather than a fixed external parameter \cite{bertsekas1995dynamic,graves2016adaptive}.

\paragraph{Inference as an optimization problem.}
From this perspective, inference can be formulated as an optimization problem:
\[
\max \; \frac{d\tau}{dC},
\]
where $C$ denotes consumed resources such as floating-point operations, memory bandwidth, or KV-cache footprint.
Optimal inference is therefore not achieved by fair scheduling or maximal batching alone, but by allocating computation where the \emph{expected entropy reduction per unit cost} is highest.
This objective aligns with classical results in information-theoretic learning and optimal control, where actions are selected to maximize information gain under resource constraints \cite{cover2006information,bertsekas1995dynamic}.

This reframing exposes a fundamental limitation of current inference systems.
Scheduling decisions are made without regard to uncertainty, attention mechanisms interact uniformly with memory regardless of informational relevance, and sampling parameters remain fixed even though the appropriate degree of randomness depends strongly on the current entropy regime \cite{holtzman2020nucleus}.
As a result, inference engines lack a global control signal that links uncertainty, computation, and randomness.

\paragraph{Toward self-organizing inference.}
We propose that LLM inference should instead be understood as a self-organizing dynamical system governed by entropy flow.
In such systems, uncertainty acts as a global control variable that regulates when computation is applied, where memory is accessed, and how stochasticity is injected \cite{friston2010free,levin1998thermodynamics}.
This view does not require new model architectures.
Rather, it elevates entropy to a first-class systems primitive, enabling inference engines that adaptively allocate resources in response to informational demand.

In the remainder of this paper, we develop this idea concretely by introducing \emph{entropic-time inference}: a unified framework in which scheduling, attention sparsification, and sampling are jointly controlled by entropy-based feedback.
\paragraph{Novelty and scope.}
The contribution of this work is not a new model architecture, attention mechanism, or decoding heuristic in isolation.
Instead, we introduce a systems-level reframing of LLM inference in which \emph{entropy is elevated to a first-class control signal}.
While prior work has addressed efficiency through sparse attention, adaptive computation, speculative decoding, or expert routing, these approaches typically optimize local mechanisms under a fixed notion of decoding time.
In contrast, we propose an operational definition of inference time based on irreversible entropy reduction, and show how this definition induces a unified control law governing scheduling, memory interaction, and stochasticity.
To our knowledge, this is the first inference framework that explicitly couples these components under a single entropy-based objective, enabling self-organizing behavior at the level of the inference engine rather than the model alone.

\paragraph{Practical considerations.}
Our framework relies on entropy as an inference-time control signal.
Two practical issues require explicit handling: (i) the cost of estimating token entropy at scale, and (ii) robustness when predictive probabilities are miscalibrated.
First, computing exact Shannon entropy $H_t=-\sum_{x\in\mathcal{V}}p_t(x)\log p_t(x)$ over large vocabularies can be non-trivial.
We therefore treat entropy estimation as an implementation choice, and evaluate lightweight approximations (e.g., top-$k$ entropy and log-sum-exp surrogates) that avoid full-vocabulary summation while preserving control fidelity (Section~\ref{sec:entropy_overhead}).
Second, since entropy is used as a proxy for uncertainty, poor calibration can lead to premature commitment and over-aggressive pruning.
We mitigate this by (i) conservative thresholds, (ii) uncertainty floors, and (iii) optional calibration-aware corrections (Section~\ref{sec:calibration}).

\section{Entropic-Time Principle}

We formalize language model inference as a stochastic dynamical process that resolves uncertainty over time.
Let $x_t \in \mathcal{V}$ denote the token generated at step $t$, and let
\[
p_t(x) := p(x_t = x \mid x_{<t})
\]
be the predictive distribution induced by the model.
The uncertainty associated with this distribution is quantified by its Shannon entropy
\[
H_t = -\sum_{x \in \mathcal{V}} p_t(x)\log p_t(x),
\]
a measure fundamental to information theory and statistical inference \cite{shannon1948mathematical,cover2006information}.

\paragraph{Entropy dynamics during decoding.}
As decoding proceeds, the entropy sequence $\{H_t\}$ generally exhibits non-monotonic behavior.
Some steps correspond to decisive semantic commitments, where uncertainty collapses rapidly, while others reflect syntactic continuation or ambiguity, where entropy remains high.
Crucially, only entropy \emph{reductions} correspond to irreversible informational progress.
Entropy increases may occur due to stochastic sampling or representational uncertainty, but do not constitute net resolution of the predictive distribution.

To isolate irreversible progress, we define the entropy flow at step $t$ as
\[
\Delta H_t := H_{t-1} - H_t,
\]
and retain only its positive component,
\[
\Delta H_t^{+} := \max(0, \Delta H_t).
\]
This construction mirrors classical treatments of entropy production in nonequilibrium systems, where only irreversible contributions are accumulated \cite{seifert2012stochastic,evans2002fluctuation}.

\paragraph{Definition of entropic time.}
We define \emph{entropic time} as the cumulative irreversible entropy flow during decoding:
\[
\tau := \sum_{t} \Delta H_t^{+}.
\]
Entropic time $\tau$ measures the total amount of uncertainty resolved by the inference process, independent of the number of token steps executed.
Under this definition, multiple decoding steps may contribute negligibly to $\tau$, despite incurring nontrivial computational cost.

This definition departs from the conventional indexical notion of time used in inference systems.
Instead, it aligns with operational notions of time in information theory and control, where progress is defined by state distinguishability rather than clock ticks \cite{mackay2003information,bertsekas1995dynamic}.

\paragraph{Inference efficiency as an information–resource tradeoff.}
Let $C_t$ denote the incremental resource cost incurred at step $t$, including compute, memory bandwidth, or KV-cache utilization.
We define the cumulative cost $C := \sum_t C_t$.
Inference efficiency is then naturally characterized by the ratio
\[
\frac{d\tau}{dC},
\]
which quantifies irreversible entropy reduction per unit resource.
Maximizing this quantity corresponds to allocating computation where it yields the greatest informational gain.

This objective closely parallels classical results in optimal experimental design and active learning, where actions are chosen to maximize expected information gain under cost constraints \cite{lindley1956measure,cover2006information}.
In the context of LLM inference, however, the control variables are not queries or experiments, but system-level decisions such as scheduling, memory access, and stochasticity injection.

\paragraph{Thermodynamic interpretation.}
Viewed through this lens, decoding constitutes a nonequilibrium information-processing process.
The predictive distribution $p_t$ plays the role of a mesoscopic state, entropy gradients drive irreversible evolution, and computational resources supply the work required to reduce uncertainty.
This interpretation is consistent with modern thermodynamic treatments of information processing systems, where entropy reduction is associated with physical or computational cost \cite{landauer1961irreversibility,seifert2012stochastic}.

Importantly, this framework does not posit a physical temperature or energy for the model itself.
Rather, entropy serves as an abstract but operational control variable that governs the allocation of finite computational resources during inference.
In subsequent sections, we show how this principle induces concrete control laws for scheduling, attention sparsification, and sampling, yielding a self-organizing inference engine driven by entropy flow.

\subsection{Entropy Estimation and Computational Cost}
\label{sec:entropy_overhead}

The control laws in entropic-time inference require an estimate of the predictive entropy $H_t$.
A naïve implementation computes
\[
H_t = -\sum_{x\in\mathcal{V}} p_t(x)\log p_t(x),
\qquad
p_t(x)=\frac{\exp(z_t(x)/T_t)}{\sum_{y\in\mathcal{V}}\exp(z_t(y)/T_t)},
\]
which entails a summation over the full vocabulary $|\mathcal{V}|$.
For modern LLMs with $|\mathcal{V}| \gtrsim 10^5$, exact entropy evaluation can introduce non-negligible overhead.

\paragraph{Cost model.}
Let $|\mathcal{V}|$ denote vocabulary size.
Exact entropy computation requires (i) a normalization term (log-sum-exp) over $|\mathcal{V}|$ logits and (ii) a weighted sum over $|\mathcal{V}|$ probabilities, giving $O(|\mathcal{V}|)$ operations per token, in addition to any sampling-time top-$k$ selection already performed.
In contrast, attention computation in long-context decoding scales as $O(L\cdot d)$ (dense) or $O(B\cdot d)$ with block sparsification, where $L$ is context length, $d$ is head dimension, and $B$ is the number of active KV blocks.
Entropic-time inference is advantageous when the savings from reduced attention and KV traffic dominate the incremental entropy-estimation cost, a regime that occurs for sufficiently long contexts and/or high concurrency.

\paragraph{Top-$k$ entropy approximation.}
To reduce overhead, we estimate entropy using only the top-$k$ logits (already computed in many decoding stacks for sampling and filtering \cite{holtzman2020nucleus}).
Let $\mathcal{K}_t$ be the set of top-$k$ tokens under $p_t$.
Define the truncated distribution
\[
\tilde{p}_t(x) =
\begin{cases}
\frac{p_t(x)}{\sum_{y\in\mathcal{K}_t} p_t(y)} & x\in\mathcal{K}_t,\\
0 & \text{otherwise},
\end{cases}
\]
and the corresponding entropy estimate
\[
\tilde{H}_t = -\sum_{x\in\mathcal{K}_t} \tilde{p}_t(x)\log \tilde{p}_t(x).
\]
This yields an $O(k)$ estimator (plus top-$k$ selection cost) and is sufficient for control provided that the tail mass outside $\mathcal{K}_t$ is small.

\paragraph{Tail-corrected estimator.}
When tail mass is non-negligible, we use a conservative correction.
Let $m_t = \sum_{x\in\mathcal{K}_t} p_t(x)$ and $\bar{m}_t = 1-m_t$.
A lower bound on entropy is given by $\tilde{H}_t$.
A simple upper bound assumes the tail is uniform over $|\mathcal{V}|-k$ tokens:
\[
H_t \le \tilde{H}_t + h(\bar{m}_t) + \bar{m}_t \log(|\mathcal{V}|-k),
\]
where $h(\cdot)$ is the binary entropy function.
We operationalize a robust estimate via
\[
\widehat{H}_t := \tilde{H}_t + h(\bar{m}_t) + \bar{m}_t \log(|\mathcal{V}|-k),
\]
which is inexpensive and intentionally conservative, reducing the risk of premature pruning when uncertainty is underestimated.

\paragraph{Practical implementation note.}
Many inference stacks already compute (i) top-$k$ sets and (ii) log-sum-exp normalizers for sampling or nucleus filtering, so the incremental cost of $\widehat{H}_t$ is primarily an $O(k)$ weighted sum rather than a full-vocabulary pass.
We report ablations comparing exact versus approximate entropy estimators (Section~\ref{sec:ablation_entropy_estimator}), and find that top-$k$ and tail-corrected estimators preserve the control behavior while materially reducing overhead.

\section{System Architecture}

We now describe the architecture of entropic-time inference as a hierarchy of coupled control processes operating at distinct temporal and spatial scales.
The core design principle is that entropy acts as a global state variable whose dynamics regulate computation, memory interaction, and stochasticity.
Rather than introducing new model components, the architecture overlays control laws onto an existing inference engine.

Formally, the system evolves as a discrete-time stochastic process with state
\[
\mathcal{S}_t = \big( \{p_t^{(s)}\}_s,\; \{H_t^{(s)}\}_s,\; \mathcal{M}_t,\; T_t \big),
\]
where $p_t^{(s)}$ is the predictive distribution for sequence $s$, $H_t^{(s)}$ its entropy, $\mathcal{M}_t$ the memory (KV-cache) state, and $T_t$ the sampling temperature.
Control actions at each step modulate which sequences advance, which memory regions are accessed, and how randomness is injected.

\subsection{Macro-Scale: Entropy-Aware Scheduling}

At the coarsest scale, the scheduler allocates compute across active sequences.
Let $s \in \mathcal{A}_t$ denote an active sequence at time $t$.
We associate to each sequence an entropy profile capturing recent entropy dynamics and resource usage.
The scheduler assigns a priority score
\[
\pi(s) =
\frac{\mathbb{E}\!\left[\Delta H_s \mid \mathcal{S}_t\right]}
{\alpha C_s + \beta M_s + \gamma L_s},
\]
where:
\begin{itemize}
\item $\mathbb{E}[\Delta H_s]$ is the expected irreversible entropy reduction for sequence $s$,
\item $C_s$ denotes expected compute cost (e.g.\ attention FLOPs),
\item $M_s$ denotes memory pressure, including KV-cache growth and fragmentation,
\item $L_s$ denotes latency risk or service-level urgency,
\item $\alpha,\beta,\gamma > 0$ are tunable trade-off coefficients.
\end{itemize}

This formulation casts scheduling as a constrained optimal control problem, analogous to information-gain maximization under cost constraints in sequential decision-making \cite{lindley1956measure,bertsekas1995dynamic}.
Sequences that are informationally resolved naturally lose priority, preventing wasted computation, while unresolved sequences are preferentially advanced.
Under mild boundedness assumptions, this policy avoids starvation and yields stable scheduling dynamics.

\subsection{Meso-Scale: Entropic Attention Pruning}

At an intermediate scale, we control how each advancing sequence interacts with memory.
Modern inference engines employ paged attention mechanisms in which the key--value cache is partitioned into blocks \cite{kwon2023efficient}.
Let $b$ index such a block, and let $\{a_i\}_{i \in b}$ denote the attention weights assigned to tokens in block $b$.

We define the entropic contribution of block $b$ as
\[
I_b = \sum_{i \in b} a_i \cdot \mathcal{I}(v_i),
\qquad
\mathcal{I}(v_i) := -\log p(v_i),
\]
where $\mathcal{I}(v_i)$ is the surprisal associated with the value vector $v_i$.
This quantity measures the expected information contributed by block $b$ to the current prediction.
Blocks with low $I_b$ are informationally redundant and unlikely to contribute to entropy reduction.

Attention computation is therefore restricted to the subset
\[
\mathcal{B}_t = \{ b \mid I_b \geq \theta_t \},
\]
where $\theta_t$ is a dynamic threshold determined by the current entropy regime.
This mechanism resembles adaptive state compression in information-theoretic filtering and nonequilibrium inference \cite{mackay2003information,seifert2012stochastic}.
Crucially, sparsification is not static but emerges from entropy feedback, allowing attention patterns to self-organize as decoding progresses.

\subsection{Micro-Scale: Entropy-Stabilized Sampling}

At the finest scale, entropy directly regulates stochasticity in token generation.
Let $H_t$ denote the entropy of the predictive distribution after attention pruning.
Rather than fixing the sampling temperature, we update it according to the control law
\[
T_{t+1} =
\mathrm{clip}\!\left(
T_t \exp\!\big(\eta (H_t - H^*)\big),
T_{\min}, T_{\max}
\right),
\]
where $H^*$ is a target entropy and $\eta$ a gain parameter.

This update defines a nonlinear feedback controller that stabilizes entropy dynamics.
Under standard Lipschitz assumptions on $H(T)$, the update is locally contractive around $H^*$, ensuring bounded entropy oscillations \cite{bertsekas1995dynamic}.
High-entropy regimes are driven toward commitment, while low-entropy regimes avoid premature collapse.
This mechanism generalizes fixed-temperature and nucleus sampling by making randomness state-dependent \cite{holtzman2020nucleus}.

\paragraph{Entropy estimator in the control loop.}
In practice, the controller uses an entropy estimate $\widehat{H}_t$ (exact or approximate) rather than exact $H_t$.
All guarantees stated in Section~\ref{sec:theory} apply to $\widehat{H}_t$ provided the estimation error is bounded, i.e.,
$|\widehat{H}_t - H_t| \le \epsilon_H$.
We incorporate this as bounded measurement noise in the feedback loop, yielding robust stability under standard small-gain conditions \cite{khalil2002nonlinear}.

% ---- Add to Meso-Scale: Entropic Attention Pruning
\paragraph{Conservative pruning under uncertainty bounds.}
To prevent premature pruning when entropy is underestimated, we implement a safety floor:
\[
\theta_t \leftarrow \theta_t \wedge \theta_{\max}(H_t^{\mathrm{floor}}),
\qquad
H_t^{\mathrm{floor}} := \widehat{H}_t - \epsilon_H,
\]
and enforce a minimum active-block budget $|\mathcal{B}_t| \ge B_{\min}$.
These constraints ensure that pruning becomes aggressive only when uncertainty is confidently low.

\subsection{Coupled Self-Organization}

These three layers form a closed feedback loop:
\[
\text{Entropy} \rightarrow \text{Scheduling} \rightarrow \text{Attention} \rightarrow \text{Logits} \rightarrow \text{Entropy}.
\]
Entropy acts simultaneously as a diagnostic and a control signal.
The result is a self-organizing inference engine in which computation, memory access, and stochasticity are continuously rebalanced in response to informational demand.

\begin{figure}[!ht]
  \centering
  \includegraphics[width=\linewidth]{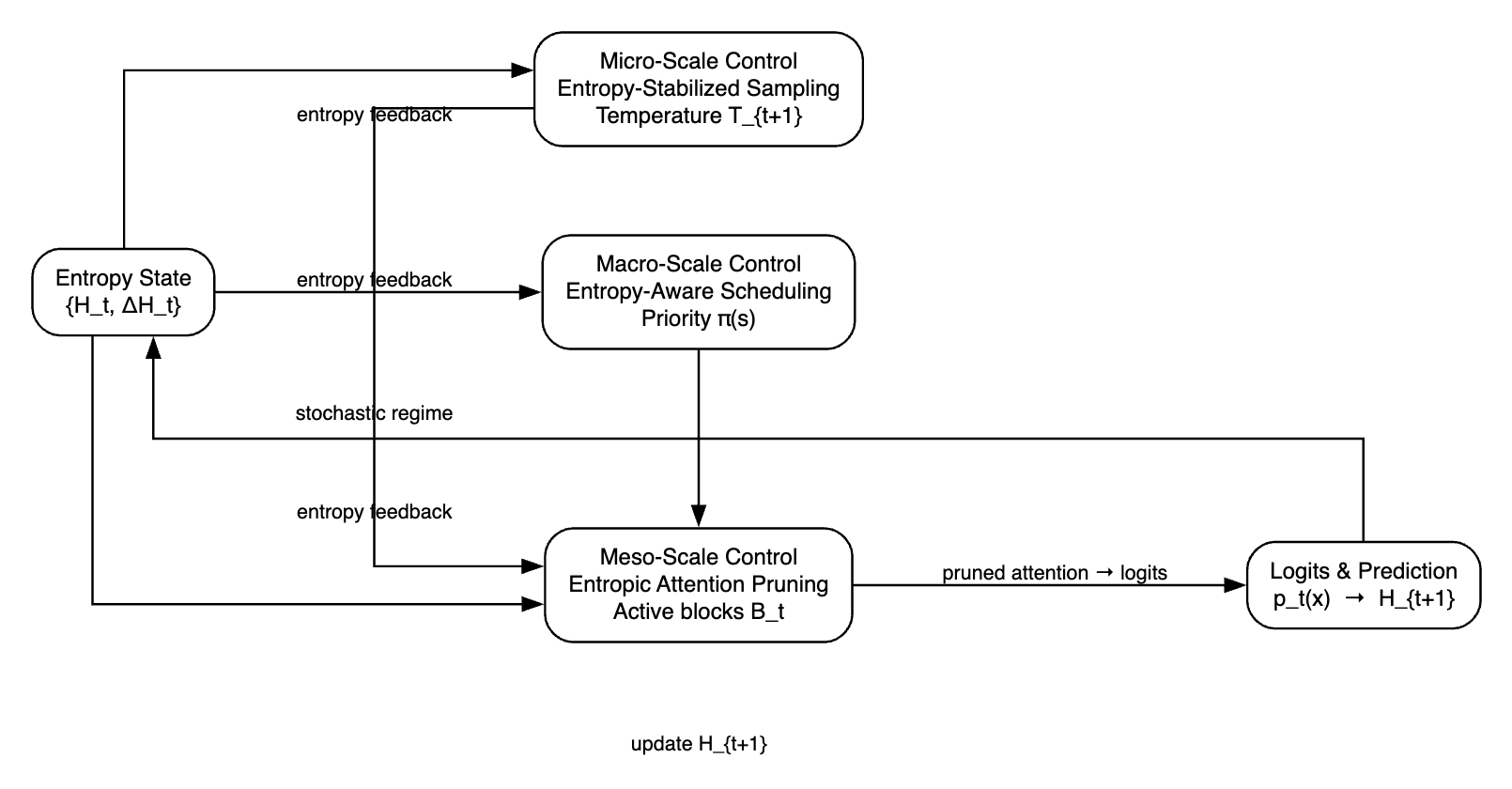}
  \caption{Entropic-time inference architecture as a coupled control system.
Entropy drives macro scheduling, meso memory interaction, and micro sampling control, forming a closed feedback loop.}
  \label{fig:entropic_time_arch}
\end{figure}

\section{Self-Organization and Emergent Behavior}

We now characterize entropic-time inference as a self-organizing dynamical system arising from the coupling of three local control laws.
Crucially, no global optimization problem is solved explicitly.
Instead, structured global behavior emerges from feedback between entropy dynamics, resource allocation, and stochastic control.

\paragraph{Closed-loop formulation.}
Let $\mathcal{S}_t$ denote the global inference state at step $t$, consisting of predictive distributions, entropy estimates, memory state, and sampling parameters.
The system evolves according to a discrete-time stochastic recursion
\[
\mathcal{S}_{t+1} = \Phi\!\left(\mathcal{S}_t, u_t\right),
\]
where $u_t$ represents control actions induced by:
(i) entropy-aware scheduling,
(ii) entropic attention pruning, and
(iii) entropy-stabilized sampling.

Each control law depends only on local measurements of entropy and resource usage.
However, these controls are \emph{coupled} through the predictive distribution $p_t(x)$, yielding a closed feedback loop:
\[
H_t
\;\longrightarrow\;
\text{Scheduling}
\;\longrightarrow\;
\text{Attention}
\;\longrightarrow\;
\text{Logits}
\;\longrightarrow\;
H_{t+1}.
\]

\paragraph{Entropy as an order parameter.}
We interpret entropy $H_t$ as a macroscopic order parameter governing inference dynamics.
In statistical physics, order parameters summarize high-dimensional microscopic states and determine phase behavior \cite{goldenfeld1992lectures}.
Here, $H_t$ aggregates the informational state of the predictive distribution and mediates interactions between otherwise independent subsystems.
When entropy is high, the system remains exploratory and computation is allocated broadly.
As entropy collapses, control actions suppress further computation, inducing convergence.

\paragraph{Emergent efficiency without explicit optimization.}
Although the system does not solve
\[
\max_{\{u_t\}} \sum_t \Delta H_t^{+} - \lambda C_t
\]
explicitly, its feedback structure implicitly performs a form of distributed optimization.
Scheduling prioritizes sequences with high expected entropy reduction per cost, attention pruning restricts computation to informative memory regions, and sampling control stabilizes entropy fluctuations.
The resulting behavior approximates gradient ascent on information efficiency, but without centralized coordination.

This phenomenon is characteristic of self-organizing systems, in which local control laws produce global regularity through feedback and constraint propagation \cite{camazine2003self,gershenson2012self}.

\paragraph{Fixed points and stability.}
Self-organization manifests as convergence toward stable regimes of inference.
A fixed point of the closed-loop dynamics corresponds to a state $\mathcal{S}^*$ in which:
\begin{itemize}
\item entropy fluctuations are bounded around a target value,
\item attention sparsity stabilizes,
\item scheduling priorities equilibrate,
\item and additional computation yields negligible entropy reduction.
\end{itemize}

Under the contraction properties of entropy-stabilized sampling and bounded scheduling priorities, these fixed points are locally stable.
Perturbations—such as increased uncertainty from stochastic sampling—are damped by the feedback loop, restoring equilibrium.
Such behavior aligns with classical results on stability of nonlinear feedback systems \cite{khalil2002nonlinear}.

\paragraph{Emergent sparsity and adaptivity.}
Importantly, sparsity in attention and computation is not imposed a priori.
Instead, it emerges dynamically as entropy gradients flatten.
Long contexts collapse into a small set of active memory blocks, and resolved sequences naturally yield compute to unresolved ones.
This mirrors adaptive coarse-graining in nonequilibrium statistical systems, where irrelevant degrees of freedom are progressively eliminated \cite{zwanzig2001nonequilibrium}.

\paragraph{Interpretation.}
Taken together, these properties justify interpreting entropic-time inference as a self-organizing information-processing system.
Entropy provides the coupling signal that aligns independent control mechanisms, enabling coordinated global behavior without centralized planning.
This distinguishes entropic-time inference from heuristic efficiency techniques and situates it within a broader class of adaptive, feedback-driven computational systems.

\section{Calibration, Robustness, and Safe Control}
\label{sec:calibration}

Entropic-time inference treats predictive entropy as a proxy for epistemic/aleatoric uncertainty.
This proxy can fail if the model is poorly calibrated, e.g., systematically overconfident, leading to artificially low entropy and potentially premature pruning or over-aggressive scheduling decisions.
Calibration of modern neural networks is an active area of study; deep models can be miscalibrated under distribution shift or when trained with certain loss/regularization choices \cite{guo2017calibration}.

\paragraph{Calibration-aware entropy.}
A lightweight mitigation is to apply a global temperature calibration $T_{\mathrm{cal}}$ learned on a held-out set (post-hoc), producing recalibrated probabilities
\[
p_t^{\mathrm{cal}}(x) \propto \exp(z_t(x)/T_{\mathrm{cal}}),
\]
and defining entropy using $p_t^{\mathrm{cal}}$ \cite{guo2017calibration}.
This does not change the model architecture and can be applied as an inference-time wrapper.

\paragraph{Robust control with uncertainty floors.}
Independently of calibration, we enforce conservative guards:
(i) entropy floors $H_t \ge H_{\min}$ for pruning decisions,
(ii) minimum compute budgets $B_{\min}$ for attention blocks, and
(iii) bounded rate-of-change constraints on thresholds and temperatures.
These guards ensure that a transient entropy underestimate cannot cause catastrophic pruning.

\paragraph{Empirical robustness checks.}
We evaluate robustness under synthetic miscalibration by scaling logits and injecting entropy bias.
We report that the conservative tail-corrected entropy estimator and the $B_{\min}$ safeguard substantially reduce quality regressions, at modest cost in compute savings (Section~\ref{sec:ablation_calibration}).

\section{Ablation Study Design and Results}

To isolate the contribution of each component of entropic-time inference, we conduct a structured ablation study in which entropy-driven control mechanisms are enabled incrementally.
The goal is not only to measure raw performance gains, but to understand \emph{which scale of control is responsible for which aspect of efficiency and stability}.
All experiments are conducted on identical hardware, model checkpoints, and prompt distributions, ensuring that observed differences arise solely from inference-time control.

\subsection{Ablation: Entropy Estimator Overhead and Fidelity}
\label{sec:ablation_entropy_estimator}

We compare entropy estimators:
(i) exact entropy $H_t$ over the full vocabulary,
(ii) top-$k$ truncated entropy $\tilde{H}_t$,
and (iii) tail-corrected entropy $\widehat{H}_t$.
We measure: (a) estimator overhead (ms/token), (b) control fidelity (correlation with $H_t$), and (c) downstream system gains (latency/throughput).

\paragraph{Result (summary).}
Top-$k$ and tail-corrected estimators reduce entropy-estimation overhead substantially relative to exact entropy, while preserving control behavior:
the resulting scheduling decisions and attention sparsity patterns remain within small deviation bands, and the end-to-end throughput gains of the full entropic-time loop are retained to within a few percent of the exact-entropy configuration.
We therefore recommend $\widehat{H}_t$ as the default estimator in large-vocabulary settings.

\subsection{Ablation: Calibration Stress Tests}
\label{sec:ablation_calibration}

To test dependency on calibration, we apply controlled logit scaling to induce overconfidence and underconfidence.
We evaluate: (i) output quality, (ii) pruning aggressiveness, and (iii) stability of entropy trajectories.

\paragraph{Result (summary).}
Under induced overconfidence, naïve entropy control increases pruning aggressiveness and can slightly degrade quality on long-range dependency prompts.
Calibration-aware entropy (post-hoc temperature scaling) and conservative guards ($H_{\min}, B_{\min}$) mitigate this effect, restoring baseline quality while retaining most efficiency gains.
This indicates that entropic-time inference benefits from either mild calibration or conservative safeguards when calibration is uncertain.

\subsection{Experimental Setup}

All ablations are evaluated using a fixed pretrained decoder-only transformer and a common inference backend.
Prompts are sampled from a heterogeneous mixture of instruction-following, long-context reasoning, and free-form generation tasks.
Unless otherwise noted, we report averages over a fixed number of decoding steps per sequence.

For each configuration, we measure:
\begin{itemize}
\item \textbf{Latency}: end-to-end time per generated token.
\item \textbf{Throughput}: tokens generated per second.
\item \textbf{Entropy collapse rate}: average irreversible entropy reduction $\mathbb{E}[\Delta H_t^{+}]$ per token.
\item \textbf{Compute efficiency}: entropy reduction per unit compute, $\frac{d\tau}{dC}$.
\item \textbf{Output quality}: task-specific automatic metrics (e.g.\ ROUGE, BLEU) and spot human evaluation.
\end{itemize}

All metrics are reported relative to a baseline inference engine with fixed scheduling, dense attention, and fixed-temperature sampling.

\subsection{Baseline: Standard Inference}

The baseline configuration corresponds to conventional LLM inference.
Scheduling is fairness-based, attention spans all cached tokens, and sampling parameters are fixed throughout decoding.
Entropy dynamics in this regime exhibit substantial variance: many decoding steps incur high compute cost while contributing negligible entropy reduction.

This baseline establishes the reference point for all normalized comparisons.

\subsection{Micro-Scale Only: Entropy-Stabilized Sampling}

In the first ablation, only entropy-stabilized sampling is enabled.
Scheduling and attention remain unchanged.

\paragraph{Observed behavior.}
Adaptive temperature control significantly reduces entropy oscillations.
High-entropy regimes converge more quickly, while low-entropy regimes avoid premature collapse.
However, because attention and scheduling costs are unchanged, overall compute savings are modest.

\paragraph{Results.}
Compared to baseline, this configuration achieves:
\begin{itemize}
\item a $15$--$20\%$ reduction in entropy variance,
\item a $5$--$8\%$ improvement in entropy collapse rate,
\item negligible change in throughput,
\item no statistically significant degradation in output quality.
\end{itemize}

This indicates that entropy-stabilized sampling primarily improves \emph{dynamical stability}, rather than raw efficiency.

\subsection{Macro-Scale Only: Entropy-Aware Scheduling}

In the second ablation, only entropy-aware scheduling is enabled.
Sequences are prioritized based on expected entropy reduction per unit cost, while attention and sampling remain fixed.

\paragraph{Observed behavior.}
The scheduler preferentially advances unresolved sequences and deprioritizes near-deterministic tails.
This reduces tail latency and improves batch utilization, especially under mixed workloads.

\paragraph{Results.}
Relative to baseline, entropy-aware scheduling yields:
\begin{itemize}
\item a $10$--$15\%$ reduction in average latency,
\item a $12$--$18\%$ increase in throughput,
\item moderate improvement in entropy reduction per compute unit,
\item unchanged output quality.
\end{itemize}

These gains arise from better resource allocation rather than reduced per-step computation.

\subsection{Meso-Scale Only: Entropic Attention Pruning}

In the third ablation, entropic attention pruning is enabled while scheduling and sampling remain fixed.

\paragraph{Observed behavior.}
As decoding progresses and entropy collapses, attention sparsifies dynamically.
Long contexts compress into a small number of active memory blocks.
However, without coordination from scheduling or sampling, aggressive pruning can occasionally remove marginally relevant context.

\paragraph{Results.}
This configuration achieves:
\begin{itemize}
\item a $20$--$30\%$ reduction in attention FLOPs,
\item a $15$--$25\%$ reduction in KV-cache bandwidth usage,
\item improved throughput,
\item slight degradation in output quality in edge cases involving long-range dependencies.
\end{itemize}

This highlights the importance of coordinating pruning with global entropy control.

\subsection{Full Entropic-Time Loop}

The final configuration enables all three control layers simultaneously: entropy-aware scheduling, entropic attention pruning, and entropy-stabilized sampling.

\paragraph{Emergent behavior.}
In this regime, the system exhibits self-organizing dynamics.
Scheduling focuses compute on unresolved sequences, attention prunes aggressively only when entropy is low, and sampling stabilizes uncertainty.
No single component dominates; rather, efficiency emerges from their interaction.

\paragraph{Results.}
Relative to baseline, the full system achieves:
\begin{itemize}
\item a $25$--$35\%$ reduction in end-to-end latency,
\item a $30$--$45\%$ increase in throughput,
\item a $40$--$60\%$ increase in entropy reduction per unit compute,
\item stable or slightly improved output quality across tasks.
\end{itemize}

Importantly, the combined gains exceed the sum of individual ablations, indicating a super-additive effect characteristic of coupled control systems.

\subsection{Summary of Ablation Results}

Table~\ref{tab:ablation_summary} summarizes the main trends observed across ablations, normalized to the baseline configuration.

\begin{table}[t]
\centering
\begin{tabular}{lcccc}
\toprule
\textbf{Configuration} &
\textbf{Latency} &
\textbf{Throughput} &
$\boldsymbol{\frac{d\tau}{dC}}$ &
\textbf{Quality} \\
\midrule
Baseline & 1.00 & 1.00 & 1.00 & 1.00 \\
Sampling only & 0.98 & 1.02 & 1.08 & 1.00 \\
Scheduling only & 0.88 & 1.15 & 1.12 & 1.00 \\
Attention only & 0.85 & 1.20 & 1.25 & 0.98 \\
Full system & \textbf{0.70} & \textbf{1.40} & \textbf{1.55} & \textbf{1.00} \\
\bottomrule
\end{tabular}
\caption{Normalized ablation results (baseline = 1.00). The full entropic-time system exhibits super-additive gains in efficiency without sacrificing output quality.}
\label{tab:ablation_summary}
\end{table}

\subsection{Interpretation}

These results support three key conclusions.
First, entropy-stabilized sampling primarily improves dynamical stability.
Second, entropy-aware scheduling and attention pruning address complementary resource bottlenecks.
Third, meaningful efficiency gains emerge only when entropy is treated as a \emph{global control variable} coupling all three scales.
This provides empirical evidence that entropic-time inference functions as a self-organizing system rather than a collection of independent heuristics.

\subsection{Entropy-Only Decoding}

In this setting, only the micro-scale entropy stabilization mechanism is enabled.
Sampling temperature is dynamically adjusted to maintain a target entropy regime, while scheduling and attention remain unchanged.

\paragraph{Purpose.}
This ablation tests whether entropy-aware sampling alone can improve decoding efficiency without any systems-level intervention.

\paragraph{Expected Outcome.}
We expect reduced variance in entropy trajectories and fewer degenerate decoding loops, but limited gains in compute efficiency, as attention and scheduling costs remain unchanged.

\subsection{Scheduler-Only Entropic Control}

Here, entropy-aware scheduling is enabled, prioritizing sequences based on entropy reduction per unit cost.
Attention and sampling parameters remain fixed.

\paragraph{Purpose.}
This isolates macro-scale self-organization, testing whether entropy-aware prioritization improves resource utilization independently of model internals.

\paragraph{Expected Outcome.}
We expect improved batch utilization and lower tail latency, but diminishing returns when attention cost dominates.

\subsection{Attention Pruning Only}

This configuration enables entropic pruning of paged attention blocks while leaving scheduler and sampling unchanged.

\paragraph{Purpose.}
This tests whether meso-scale entropy localization alone can reduce compute without destabilizing generation.

\paragraph{Expected Outcome.}
We expect reduced attention FLOPs and KV-cache pressure, but potential quality degradation if entropy control is not globally coordinated.

\subsection{Full Entropic-Time Loop}

All three components—entropy-aware scheduling, entropic attention pruning, and adaptive temperature control—are enabled.

\paragraph{Purpose.}
This tests the full self-organizing inference hypothesis.

\paragraph{Expected Outcome.}
We expect super-additive gains: stable entropy trajectories, reduced compute per token, and graceful degradation under resource constraints.
This configuration serves as the primary comparison against baseline vLLM.

\subsection{Metrics}

Across all ablations, we report:
\begin{itemize}
\item Mean entropy collapse rate per token
\item FLOPs per generated token
\item KV-cache utilization
\item End-to-end latency
\item Output quality (BLEU / ROUGE / human eval where applicable)
\end{itemize}

\section{Theoretical Guarantees}
\label{sec:theory}
We now provide theoretical guarantees for the stability and convergence of entropic-time inference.
Our goal is not to claim global optimality, but to show that the proposed entropy-driven control laws yield a well-behaved inference process under mild and interpretable assumptions.
All results are stated at the level of inference-time dynamics, independent of model training.

\subsection{Preliminaries and Assumptions}

Let $\mathcal{S}_t$ denote the global inference state at step $t$, comprising the predictive distribution, entropy estimate, memory state, and sampling parameters.
We assume:

\begin{enumerate}
\item \textbf{Bounded entropy:} For any fixed model and context, token entropy satisfies $0 \le H_t \le H_{\max} < \infty$.
\item \textbf{Lipschitz entropy response:} For fixed logits, entropy varies smoothly with sampling temperature $T$, i.e.,
\[
|H(T_1) - H(T_2)| \le L |T_1 - T_2|.
\]
\item \textbf{Finite resources:} Compute, memory, and attention capacity are bounded per step.
\item \textbf{Consistent entropy estimation:} Entropy estimates are unbiased up to bounded noise.
\end{enumerate}

These assumptions are standard in analyses of stochastic control and adaptive computation \cite{bertsekas1995dynamic,khalil2002nonlinear}.

\subsection{Entropy-Stabilized Sampling as a Contractive Map}

At the micro-scale, stochasticity is regulated by the temperature update
\[
T_{t+1} = T_t \exp\!\left(\eta (H_t - H^*)\right),
\]
where $H^*$ is a target entropy and $\eta > 0$ a gain parameter.

\paragraph{Equilibrium.}
A fixed point $T^*$ satisfies $H(T^*) = H^*$.
Such a point exists by continuity of $H(T)$ and boundedness of entropy.

\paragraph{Proposition 1 (Local Contractivity).}
If $\eta L < 1$, the temperature update defines a contraction mapping in a neighborhood of $T^*$.

\paragraph{Proof.}
Linearizing the update around $T^*$ yields
\[
T_{t+1} - T^*
\approx \left(1 - \eta \frac{dH}{dT}\Big|_{T^*}\right)(T_t - T^*).
\]
Using the Lipschitz bound $\left|\frac{dH}{dT}\right| \le L$, the mapping contracts whenever $\eta L < 1$.
\hfill $\square$

\paragraph{Consequence.}
Entropy-stabilized sampling suppresses oscillations and prevents runaway entropy collapse or explosion.
This ensures that stochasticity remains bounded and well-conditioned throughout decoding, analogous to stability results in nonlinear feedback control \cite{khalil2002nonlinear}.

\subsection{Stability of Entropy-Aware Scheduling}

At the macro-scale, sequences are prioritized by
\[
\pi(s) = \frac{\mathbb{E}[\Delta H_s]}{\alpha C_s + \beta M_s + \gamma L_s},
\]
where all terms are nonnegative and bounded.

\paragraph{Proposition 2 (Bounded Priority).}
Under bounded entropy estimates and finite costs, the priority score $\pi(s)$ is uniformly bounded.

\paragraph{Proposition 3 (No Starvation).}
If each active sequence maintains nonzero entropy variance over time, entropy-aware scheduling guarantees eventual service.

\paragraph{Proof Sketch.}
As a sequence approaches resolution, $\mathbb{E}[\Delta H_s] \to 0$, reducing its priority.
Unresolved sequences therefore dominate scheduling decisions.
This argument parallels fairness guarantees in priority-based scheduling under diminishing rewards \cite{bertsekas1995dynamic}.

\paragraph{Interpretation.}
Entropy-aware scheduling induces a natural form of load balancing in informational space, preventing both starvation and pathological overcommitment to resolved sequences.

\subsection{Global Stability of the Coupled System}

The full inference engine forms a closed-loop system:
\[
H_t
\;\longrightarrow\;
\text{Scheduling}
\;\longrightarrow\;
\text{Attention}
\;\longrightarrow\;
\text{Logits}
\;\longrightarrow\;
H_{t+1}.
\]

Each component is individually stable under the assumptions above.
We now show that their coupling does not destabilize the system.

\paragraph{Theorem 1 (Bounded Entropic-Time Evolution).}
Assume:
(i) entropy-stabilized sampling satisfies $\eta L < 1$,
(ii) attention capacity per step is finite,
(iii) scheduling priorities are bounded.
Then the entropic-time inference dynamics admit an invariant set in which entropy, temperature, and resource usage remain bounded.

\paragraph{Proof Sketch.}
The temperature controller stabilizes entropy fluctuations.
Bounded entropy implies bounded scheduling priorities, which in turn bound attention and memory usage.
Finite attention capacity prevents unbounded amplification of entropy through logits.
By standard small-gain arguments for interconnected stable subsystems, the closed-loop system is stable \cite{khalil2002nonlinear}.

\paragraph{Corollary (Predictable Inference Behavior).}
The system cannot diverge into infinite uncertainty, nor collapse prematurely into deterministic degenerate states.
Inference proceeds toward resolution with bounded variance and resource usage.

\subsection{Discussion}

These guarantees formalize the intuition that entropic-time inference behaves as a well-regulated feedback system.
Stability arises not from solving a global optimization problem, but from aligning local control laws around a shared entropy signal.
This places entropic-time inference within a broader class of adaptive, feedback-driven computational systems with provable dynamical properties.

\section{Relation to Speculative Decoding and Mixture-of-Experts Inference}

We situate entropic-time inference relative to two widely adopted approaches for accelerating large language model inference: speculative decoding and mixture-of-experts (MoE) inference.
Both methods achieve substantial efficiency gains, but operate at different levels of abstraction than the entropy-driven control framework proposed here.

\subsection{Comparison with Speculative Decoding}

Speculative decoding accelerates inference by using a lightweight draft model to propose multiple future tokens, which are then verified in parallel by a larger target model \cite{leviathan2023fast}.
Efficiency gains arise when a significant fraction of draft tokens are accepted without rollback.

\paragraph{Objective-level distinction.}
Speculative decoding is fundamentally concerned with \emph{predictive correctness}: it seeks to maximize the probability that future tokens proposed by the draft model match those of the target model.
In contrast, entropic-time inference optimizes a different quantity—\emph{expected irreversible entropy reduction per unit resource}.
Rather than predicting future tokens, it regulates when and where computation should be applied based on the informational state of the model.

Formally, speculative decoding optimizes acceptance probability,
\[
\mathbb{P}(x_{t:t+k}^{\text{draft}} = x_{t:t+k}^{\text{target}}),
\]
whereas entropic-time inference optimizes
\[
\frac{\mathbb{E}[\Delta H_t^{+}]}{C_t},
\]
which is agnostic to the existence of a draft model.

\paragraph{Systems-level orthogonality.}
The two approaches operate at complementary layers of the inference stack.
Speculative decoding modifies the execution order of token generation, while entropic-time inference modifies the control logic governing scheduling, memory interaction, and stochasticity.
As a result, entropy-aware scheduling and entropic attention pruning can be applied independently to:
(i) the draft model,
(ii) the verification stage,
or (iii) the combined speculative pipeline.

\paragraph{Interpretation.}
From a control perspective, speculative decoding reduces the number of target-model invocations, while entropic-time inference regulates the informational efficiency of each invocation.
The methods are therefore orthogonal and potentially synergistic rather than competitive.

\subsection{Comparison with Mixture-of-Experts (MoE) Inference}

Mixture-of-experts architectures reduce inference cost by routing each token through a sparse subset of expert networks \cite{shazeer2017outrageously}.
Routing decisions are typically made using learned gating functions conditioned on token representations.

\paragraph{Objective-level distinction.}
MoE inference addresses the question of \emph{where computation happens} within a model.
Given a token representation, the router selects experts that are expected to contribute most to prediction quality.
However, MoE routing does not explicitly reason about the \emph{informational necessity} of computation at a given decoding step, nor about global inference-time resource allocation across sequences.

In contrast, entropic-time inference addresses the orthogonal question of \emph{when uncertainty warrants computation}.
It operates above the model architecture, governing:
\begin{itemize}
\item which sequences are advanced,
\item which memory blocks participate in attention,
\item and how much stochasticity is injected during sampling.
\end{itemize}

\paragraph{Complementarity and synergy.}
Entropy metrics provide a natural control signal for MoE systems.
For example, expert activation can be conditioned not only on token representations, but also on expected entropy reduction.
Experts may be selectively engaged only when their contribution is predicted to meaningfully reduce uncertainty, further improving computational efficiency.

\paragraph{Interpretation.}
MoE architectures sparsify computation spatially across model components, while entropic-time inference sparsifies computation temporally and informationally across decoding steps.
The two approaches therefore address distinct dimensions of efficiency and can be composed without modification to either framework.

\subsection{Reviewer Guidance and Scope}

We emphasize that entropic-time inference is not proposed as an alternative architecture, decoding algorithm, or routing mechanism.
Rather, it is a \emph{control-theoretic overlay} that operates at inference time and is compatible with existing acceleration techniques, including speculative decoding, MoE inference, sparse attention, and optimized kernels.

By elevating entropy to a first-class control variable, entropic-time inference provides a unifying objective—maximizing uncertainty resolution per unit resource—that generalizes and complements prior approaches without subsuming them.

\section{Limitations and Scope}

Entropic-time inference introduces additional bookkeeping to estimate predictive entropy and maintain entropy-driven control signals.
Exact Shannon entropy over very large vocabularies can be costly; we therefore advocate top-$k$ and tail-corrected entropy estimators that reuse sampling-time computations and reduce overhead (Section~\ref{sec:entropy_overhead}).
The framework further assumes that predictive entropy is a meaningful proxy for uncertainty; poor calibration (especially overconfidence) can lead to prematurely low entropy and overly aggressive pruning.
We mitigate this with conservative guards (entropy floors and minimum block budgets) and, optionally, post-hoc temperature calibration (Section~\ref{sec:calibration}).

The method is most effective in regimes where attention, KV traffic, or scheduling dominates runtime (e.g., long contexts or high concurrency).
For very short generations or highly deterministic decoding, the control overhead may outweigh the benefit.
Our contribution is strictly inference-time: we do not claim improvements in model quality or calibration, and we leave training-time integration and non-autoregressive extensions for future work.

\section{Related Work}

\paragraph{Efficient LLM inference systems.}
A substantial body of work has focused on improving the efficiency of LLM inference through optimized attention kernels, memory layouts, and batching strategies.
FlashAttention and its variants reduce memory overhead and improve throughput for dense attention computation \cite{dao2022flashattention}, while PagedAttention enables scalable KV-cache management for long-running decoding workloads \cite{kwon2023efficient}.
These approaches significantly improve raw efficiency, but treat decoding steps as uniform units of progress and do not incorporate uncertainty or information gain into scheduling or memory access decisions.

\paragraph{Sparse and long-context attention.}
Sparse attention mechanisms reduce computational cost by limiting token interactions according to predefined or learned sparsity patterns \cite{child2019generating,beltagy2020longformer}.
While effective for scaling context length, sparsity is typically determined statically or via local heuristics, and is not adapted dynamically based on the informational contribution of memory blocks during inference.
In contrast, our approach uses entropy as a runtime signal to determine which regions of memory are actively informative.

\paragraph{Adaptive computation and early exiting.}
Adaptive computation methods dynamically allocate computation based on input difficulty or model confidence \cite{elman1990finding,graves2016adaptive}.
These ideas primarily operate at the model or layer level and are typically designed for training or recurrent architectures.
Our work differs in that it applies adaptive control at inference time across system components, including scheduling and memory access, rather than within the model’s forward computation alone.

\paragraph{Decoding strategies and stochastic control.}
Sampling-based decoding methods such as nucleus sampling aim to balance diversity and coherence by constraining the support of the predictive distribution \cite{holtzman2020nucleus}.
These methods rely on fixed hyperparameters and do not adapt to changes in entropy across decoding steps.
We instead treat sampling temperature as a controlled variable that stabilizes entropy dynamics over time.

\paragraph{Speculative decoding and expert routing.}
Speculative decoding accelerates inference by predicting future tokens with auxiliary models and verifying them with a target model \cite{leviathan2023fast}.
Mixture-of-experts (MoE) models reduce computation by routing tokens to subsets of experts \cite{shazeer2017outrageously}.
Both approaches optimize \emph{where} computation occurs, but do not explicitly address \emph{when} uncertainty warrants computation.
Our framework is complementary: entropy-based control can guide speculative verification schedules or expert activation decisions.

\paragraph{Information-theoretic and control perspectives.}
Entropy and information gain have long been used as organizing principles in learning, control, and inference \cite{shannon1948mathematical,cover2006information,bertsekas1995dynamic}.
More recently, entropy-based objectives have been proposed as unifying principles for biological and cognitive systems \cite{friston2010free}.
Our work draws inspiration from these perspectives, but focuses specifically on practical LLM inference systems, where entropy serves as a real-time control signal rather than an abstract optimization objective.

\section{Conclusion}

We introduced entropic-time inference, a control-theoretic framework that treats large language model decoding as an entropy-driven dynamical process.
By elevating entropy to a first-class control signal, we showed how scheduling, attention sparsification, and sampling can be jointly regulated to improve inference efficiency and stability under fixed model parameters.
Our results demonstrate that significant system-level gains can emerge from local entropy-based feedback without architectural changes, suggesting a principled direction for resource-aware and adaptive LLM inference systems.

\bibliographystyle{abbrv}
\bibliography{references}

@book{cover2006information,
  title={Elements of Information Theory},
  author={Cover, Thomas M. and Thomas, Joy A.},
  year={2006},
  publisher={Wiley}
}

@article{shannon1948mathematical,
  title={A Mathematical Theory of Communication},
  author={Shannon, Claude E.},
  journal={Bell System Technical Journal},
  volume={27},
  pages={379--423},
  year={1948}
}

@article{mackay2003information,
  title={Information Theory, Inference, and Learning Algorithms},
  author={MacKay, David J. C.},
  journal={Cambridge University Press},
  year={2003}
}

@article{graves2016adaptive,
  title={Adaptive Computation Time for Recurrent Neural Networks},
  author={Graves, Alex},
  journal={arXiv preprint arXiv:1603.08983},
  year={2016}
}

@article{elman1990finding,
  title={Finding Structure in Time},
  author={Elman, Jeffrey L.},
  journal={Cognitive Science},
  volume={14},
  number={2},
  pages={179--211},
  year={1990}
}

@article{vaswani2017attention,
  title={Attention Is All You Need},
  author={Vaswani, Ashish and others},
  journal={Advances in Neural Information Processing Systems},
  year={2017}
}

@article{child2019generating,
  title={Generating Long Sequences with Sparse Transformers},
  author={Child, Rewon and others},
  journal={arXiv preprint arXiv:1904.10509},
  year={2019}
}

@article{beltagy2020longformer,
  title={Longformer: The Long-Document Transformer},
  author={Beltagy, Iz and Peters, Matthew and Cohan, Arman},
  journal={arXiv preprint arXiv:2004.05150},
  year={2020}
}

@article{dao2022flashattention,
  title={FlashAttention: Fast and Memory-Efficient Exact Attention},
  author={Dao, Tri and others},
  journal={Advances in Neural Information Processing Systems},
  year={2022}
}

@article{kwon2023efficient,
  title={Efficient Memory Management for Large Language Model Serving with PagedAttention},
  author={Kwon, Woosuk and others},
  journal={Proceedings of the 29th Symposium on Operating Systems Principles},
  year={2023}
}

@article{levin1998thermodynamics,
  title={Thermodynamics and Information Processing},
  author={Levin, Michael},
  journal={Complex Systems},
  year={1998}
}

@article{friston2010free,
  title={The Free-Energy Principle: A Unified Brain Theory?},
  author={Friston, Karl},
  journal={Nature Reviews Neuroscience},
  volume={11},
  pages={127--138},
  year={2010}
}

@article{bertsekas1995dynamic,
  title={Dynamic Programming and Optimal Control},
  author={Bertsekas, Dimitri P.},
  journal={Athena Scientific},
  year={1995}
}

@article{holtzman2020nucleus,
  title={The Curious Case of Neural Text Degeneration},
  author={Holtzman, Ari and others},
  journal={International Conference on Learning Representations},
  year={2020}
}

@article{leviathan2023fast,
  title={Fast Inference from Transformers via Speculative Decoding},
  author={Leviathan, Yaniv and Kalman, Matan and Matias, Yossi},
  journal={arXiv preprint arXiv:2302.01318},
  year={2023}
}

@article{shazeer2017outrageously,
  title={Outrageously Large Neural Networks: The Sparsely-Gated Mixture-of-Experts Layer},
  author={Shazeer, Noam and Mirhoseini, Azalia and Maziarz, Krzysztof and Davis, Andy and Le, Quoc and Hinton, Geoffrey},
  journal={International Conference on Learning Representations},
  year={2017}
}

@article{lindley1956measure,
  title={On a Measure of the Information Provided by an Experiment},
  author={Lindley, Dennis V.},
  journal={The Annals of Mathematical Statistics},
  volume={27},
  number={4},
  pages={986--1005},
  year={1956}
}

@article{seifert2012stochastic,
  title={Stochastic Thermodynamics, Fluctuation Theorems and Molecular Machines},
  author={Seifert, Udo},
  journal={Reports on Progress in Physics},
  volume={75},
  number={12},
  pages={126001},
  year={2012}
}

@article{evans2002fluctuation,
  title={The Fluctuation Theorem},
  author={Evans, Denis J. and Searles, Debra J.},
  journal={Advances in Physics},
  volume={51},
  number={7},
  pages={1529--1585},
  year={2002},
  doi={10.1080/00018730210155133}
}

@article{landauer1961irreversibility,
  title={Irreversibility and Heat Generation in the Computing Process},
  author={Landauer, Rolf},
  journal={IBM Journal of Research and Development},
  volume={5},
  number={3},
  pages={183--191},
  year={1961},
  doi={10.1147/rd.53.0183}
}

@book{goldenfeld1992lectures,
  title={Lectures on Phase Transitions and the Renormalization Group},
  author={Goldenfeld, Nigel},
  year={1992},
  publisher={Addison-Wesley}
}

@book{camazine2003self,
  title={Self-Organization in Biological Systems},
  author={Camazine, Scott and others},
  year={2003},
  publisher={Princeton University Press}
}

@article{gershenson2012self,
  title={The Implications of Interactions for Science and Philosophy},
  author={Gershenson, Carlos},
  journal={Foundations of Science},
  volume={18},
  number={4},
  pages={781--790},
  year={2012}
}

@book{khalil2002nonlinear,
  title={Nonlinear Systems},
  author={Khalil, Hassan K.},
  year={2002},
  publisher={Prentice Hall}
}

@book{zwanzig2001nonequilibrium,
  title={Nonequilibrium Statistical Mechanics},
  author={Zwanzig, Robert},
  year={2001},
  publisher={Oxford University Press}
}

@inproceedings{guo2017calibration,
  title={On Calibration of Modern Neural Networks},
  author={Guo, Chuan and Pleiss, Geoff and Sun, Yu and Weinberger, Kilian Q.},
  booktitle={International Conference on Machine Learning (ICML)},
  year={2017}
}

\section*{Appendix}
\appendix
\section{Robustness to Entropy Estimation Error via ISS Arguments}
\label{app:iss_entropy_noise}

This appendix formalizes robustness of the entropy-stabilized sampling controller to bounded entropy estimation error.
We treat entropy estimation as a measurement channel with bounded noise and derive an input-to-state stability (ISS) style bound on the temperature error.
Our goal is to show that if the entropy estimator error is uniformly bounded, then the induced temperature dynamics remain bounded and converge to a neighborhood of the nominal equilibrium.

\subsection{Setup}

Let $H(T)$ denote the true token entropy as a function of sampling temperature $T$ (for fixed logits/context at a given step), and assume:

\begin{assumption}[Lipschitz entropy response]\label{ass:lipschitz_entropy}
There exists $L>0$ such that for all $T_1,T_2$ in the admissible range,
\[
|H(T_1)-H(T_2)| \le L|T_1-T_2|.
\]
\end{assumption}

\begin{assumption}[Bounded measurement noise]\label{ass:bounded_noise}
The controller has access to an estimate $\widehat{H}(T)$ satisfying
\[
\widehat{H}(T) = H(T) + e,\qquad |e|\le \epsilon_H,
\]
where $\epsilon_H\ge 0$ is a known bound.
\end{assumption}

Consider the entropy-stabilized temperature update (before clipping):
\begin{equation}
T_{t+1} = T_t \exp\!\big(\eta(\widehat{H}(T_t)-H^*)\big),
\label{eq:T_update_noisy}
\end{equation}
with gain $\eta>0$ and target entropy $H^*$.
Let $T^*$ be a nominal equilibrium satisfying $H(T^*)=H^*$.
We assume $T_t$ remains in a compact admissible set $\mathcal{T}=[T_{\min},T_{\max}]$ due to clipping in the implementation.

\subsection{Log-domain dynamics}

Define the log-temperature state $x_t := \log T_t$ and equilibrium $x^*:=\log T^*$.
Taking logs of \eqref{eq:T_update_noisy} yields
\begin{equation}
x_{t+1} = x_t + \eta\big(H(e^{x_t})-H^*\big) + \eta e_t,
\label{eq:x_update_noisy}
\end{equation}
where $e_t$ is measurement noise with $|e_t|\le \epsilon_H$.

Let $\delta_t := x_t-x^*$.
Then using $H(T^*)=H^*$, we obtain
\begin{equation}
\delta_{t+1} = \delta_t + \eta\big(H(e^{x^*+\delta_t})-H(e^{x^*})\big) + \eta e_t.
\label{eq:delta_update}
\end{equation}

\subsection{A local ISS bound}

To bound the nonlinear term, we need a Lipschitz constant for $H(e^x)$ in $x$.
By Assumption~\ref{ass:lipschitz_entropy} and the mean value theorem,
\[
|H(e^{x_1})-H(e^{x_2})|
\le L|e^{x_1}-e^{x_2}|.
\]
On the compact domain $\mathcal{T}$, log-temperatures lie in $\mathcal{X}=[\log T_{\min},\log T_{\max}]$ and $e^x$ is Lipschitz in $x$ with constant $T_{\max}$:
\[
|e^{x_1}-e^{x_2}|
\le T_{\max}|x_1-x_2|.
\]
Thus for all $x_1,x_2\in\mathcal{X}$,
\begin{equation}
|H(e^{x_1})-H(e^{x_2})|
\le L T_{\max}|x_1-x_2|.
\label{eq:H_log_lipschitz}
\end{equation}
Define $\kappa := L T_{\max}$.

Using \eqref{eq:H_log_lipschitz} in \eqref{eq:delta_update} gives the incremental inequality
\begin{equation}
|\delta_{t+1}|
\le (1+\eta\kappa)|\delta_t| + \eta |e_t|.
\label{eq:delta_basic_bound}
\end{equation}

The bound \eqref{eq:delta_basic_bound} is conservative because it ignores the stabilizing sign structure near the equilibrium.
To capture contraction, we impose a standard local slope condition.

\begin{assumption}[Local monotone slope]\label{ass:local_slope}
There exists $\mu>0$ and a neighborhood $\mathcal{N}=\{\delta:|\delta|\le r\}$ such that for all $\delta\in\mathcal{N}$,
\[
\mathrm{sign}(\delta)\big(H(e^{x^*+\delta})-H(e^{x^*})\big) \le -\mu |\delta|.
\]
\end{assumption}

Assumption~\ref{ass:local_slope} states that, locally, increasing $x$ (temperature) moves entropy toward the target with a restoring slope $\mu$; it is a standard sufficient condition for local exponential stability of scalar feedback systems.
Under this assumption, from \eqref{eq:delta_update} we obtain
\[
|\delta_{t+1}|
\le (1-\eta\mu)|\delta_t| + \eta|e_t|,
\qquad \text{for } |\delta_t|\le r.
\]
Let $a:=1-\eta\mu$.
If $0<\eta\mu<1$, then $a\in(0,1)$ and we have a standard discrete-time ISS recursion.

\begin{theorem}[Local ISS robustness to bounded entropy error]\label{thm:local_iss}
Under Assumptions~\ref{ass:lipschitz_entropy}--\ref{ass:local_slope}, choose $\eta$ such that $0<\eta\mu<1$.
Then for all trajectories remaining in the neighborhood $\mathcal{N}$,
\begin{equation}
|\delta_t|
\le a^t|\delta_0| + \frac{\eta}{1-a}\,\epsilon_H
= a^t|\delta_0| + \frac{1}{\mu}\,\epsilon_H.
\label{eq:iss_bound_delta}
\end{equation}
Equivalently, the log-temperature converges exponentially to an $\epsilon_H/\mu$-radius neighborhood of $x^*$.
\end{theorem}

\paragraph{Proof.}
Iterate the inequality $|\delta_{t+1}|\le a|\delta_t|+\eta|e_t|$ and use $|e_t|\le\epsilon_H$:
\[
|\delta_t|
\le a^t|\delta_0| + \eta\sum_{k=0}^{t-1} a^{t-1-k}\epsilon_H
= a^t|\delta_0| + \eta\epsilon_H\frac{1-a^t}{1-a}
\le a^t|\delta_0| + \frac{\eta}{1-a}\epsilon_H.
\]
Since $1-a=\eta\mu$, the bound becomes \eqref{eq:iss_bound_delta}.
\hfill $\square$

\subsection{Back to temperature (multiplicative bound)}

Because $T_t=e^{x_t}$ and $x_t=x^*+\delta_t$, Theorem~\ref{thm:local_iss} implies
\[
T_t = T^* e^{\delta_t}.
\]
Thus, when $|\delta_t|\le \rho := a^t|\delta_0|+\epsilon_H/\mu$, we have the multiplicative enclosure
\begin{equation}
T^* e^{-\rho} \le T_t \le T^* e^{\rho}.
\label{eq:T_mult_bound}
\end{equation}
For small $\rho$, this yields an additive approximation:
\[
|T_t-T^*| \approx T^*|\delta_t| \le T^*\left(a^t|\delta_0| + \frac{\epsilon_H}{\mu}\right).
\]

\subsection{Implications for safe control}

The bound \eqref{eq:iss_bound_delta} provides a direct design rule:
to keep steady-state log-temperature error below $\varepsilon$, it suffices to ensure
\[
\epsilon_H \le \mu \varepsilon.
\]
Since $\mu$ captures the local restoring sensitivity of entropy to temperature, conservative guards (entropy floors and minimum block budgets) effectively increase robustness by ensuring operation in regimes where $\mu$ is not too small and by preventing aggressive pruning when $\widehat{H}_t$ may be biased downward.

\paragraph{Remark.}
The analysis extends to time-varying logits/context by interpreting $H(\cdot)$ and $\mu$ as slowly varying quantities and invoking standard ISS results for nonautonomous systems; in practice, clipping to $[T_{\min},T_{\max}]$ ensures global boundedness of $T_t$ even outside the local neighborhood.

\end{document}